\documentclass[letterpaper, 10 pt, conference]{ieeeconf}  

\IEEEoverridecommandlockouts                              

\usepackage{cite}
\usepackage{amsmath,amssymb,amsfonts}
\usepackage{algorithmic}
\usepackage{algorithm}
\usepackage{graphicx}
\usepackage{textcomp}
\usepackage{xcolor}
\usepackage{multirow}
\usepackage{float}
\usepackage{amsmath}
\usepackage{bm}
\usepackage{subcaption}
\usepackage{soul}
\usepackage{url}
\usepackage{cool}
\usepackage{nameref}
\usepackage{float}
\usepackage[utf8x]{inputenc} 
\usepackage[breaklinks=true,bookmarks=false]{hyperref}
\usepackage{textcomp}

\DeclareMathOperator*{\argmax}{argmax}

\title{\LARGE \bf
RAPID-RL: A \underline{R}econfigurable \underline{A}rchitecture with \underline{P}reemptive-Ex\underline{i}ts for Efficient \underline{D}eep-\underline{R}einforcement \underline{L}earning
}

\author{Adarsh Kumar Kosta$^{1}$, Malik Aqeel Anwar$^{2}$, Priyadarshini Panda$^{3}$, Arijit Raychowdhury,$^{2}$ and Kaushik Roy$^{1}$
\thanks{$^{1}$Adarsh Kumar Kosta and Kaushik Roy are with Purdue University, West Lafayette, IN 47907, USA {\tt\small \{akosta, kaushik\}@purdue.edu}}
\thanks{$^{3}$Malik Aqeel Anwar and Arijit Raychowdhury are with Georgia Institute of Technology, Atlanta, GA 30332, USA {\tt\small \{aqeel.anwar, arijit.raychowdhury\}@gatech.edu}}
\thanks{$^{2}$Priyadarshini Panda is with Yale University, New Haven, CT 06520, USA {\tt\small \{priya.panda\}@yale.edu}}
}

\begin{document}

\newcommand{\ignore}[1]{}

\newif\ifsubmit
\submitfalse
\ifsubmit
    \newcommand{\akosta}[1]{}
    \newcommand{\kaushik}[1]{}
    \newcommand{\todo}[1]{}
    \newcommand{\tocite}[1]{}
    \newcommand{\update}[1]{}
\else
    \newcommand{\akosta}[1]{[{\color{brown}AK: #1}]}
    \newcommand{\kaushik}[1]{[{\color{cyan}KR: #1}]}
    \newcommand{\todo}[1]{[{\color{red}TODO: #1}]}
    \newcommand{\tocite}[1]{[{\color{red}CITE: #1}]}
    \newcommand{\update}[1]{{\color{brown} #1}}
\fi

\maketitle
\thispagestyle{empty}
\pagestyle{empty}

\begin{abstract}
Present-day Deep Reinforcement Learning (RL) systems show great promise towards building intelligent agents surpassing human-level performance. However, the computational complexity associated with the underlying deep neural networks (DNNs) leads to power-hungry implementations. This makes deep RL systems unsuitable for deployment on resource-constrained edge devices. To address this challenge, we propose a reconfigurable architecture with preemptive exits for efficient deep RL (RAPID-RL). RAPID-RL enables conditional activation of DNN layers based on the difficulty level of inputs. This allows to dynamically adjust the compute effort during inference while maintaining competitive performance. We achieve this by augmenting a deep Q-network (DQN) with side-branches capable of generating intermediate predictions along with an associated confidence score. We also propose a novel training methodology for learning the actions and branch confidence scores in a dynamic RL setting. Our experiments evaluate the proposed framework for Atari 2600 gaming tasks and a realistic Drone navigation task on an open-source drone simulator (PEDRA). We show that RAPID-RL incurs 0.34$\times$ (0.25$\times$) number of operations (OPS) while maintaining performance above 0.88$\times$ (0.91$\times$) on Atari (Drone navigation) tasks, compared to a baseline-DQN without any side-branches. The reduction in OPS leads to fast and efficient inference, proving to be highly beneficial for the resource-constrained edge where making quick decisions with minimal compute is essential.

\end{abstract}

\section{INTRODUCTION} \label{sec:intro}
Recent advances in Deep Reinforcement Learning (RL) have proven to be effective in several sequential decision-making and control tasks \cite{mnih2015human, schulman2015trpo}. Notable examples include Deep Q-Networks (DQN) \cite{mnih2013playing} and AlphaGo \cite{silver2017mastering} surpassing human-level performance on Atari 2600 games and the game of Go, respectively. These can be greatly accredited to the learning capability offered by Deep Neural Networks (DNNs) consisting of millions of parameters. However, these achievements come at the cost of high computing complexity and memory bandwidth requirement \cite{han2015learning, han2015deep, sze2017efficient} associated with DNNs. This hinders the deployment of such large-scale RL systems on edge devices with limited computing capabilities. 

\begin{figure}[ht]
\centering
\includegraphics[width=0.98\linewidth]{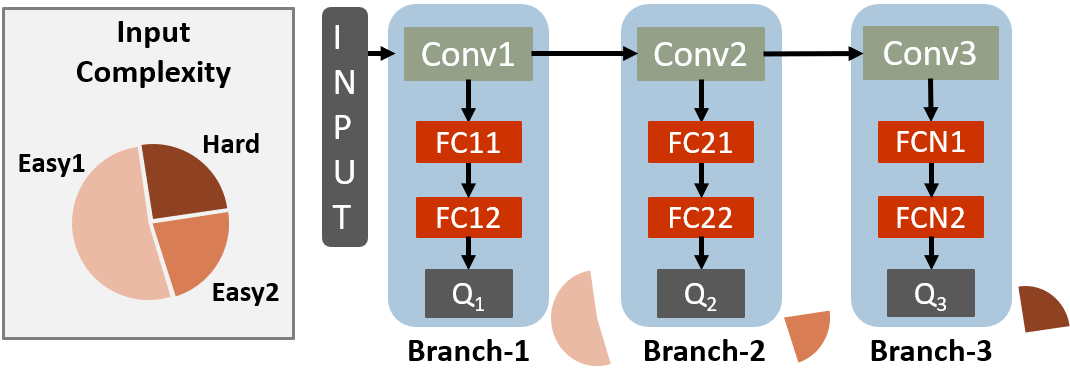}
\caption{\small{A RAPID-RL framework with two side-branches added to the baseline network consisting of three $Conv$ layers. The `easy' inputs are classified at early branches while the latter layers are activated only for `hard' inputs.}}
\label{fig:arch}
\vspace{-5mm}
\end{figure}

Deep RL systems at the edge majorly perform repeated inferences in a sustained fashion. In a real-world environment, the difficulty level of the agent inputs can vary widely. This opens up the opportunity to dynamically adjust the compute effort required to make a correct prediction based on the input difficulty itself. In such a scenario, employing DNNs with varying levels of complexity seems to be a promising approach. Smaller networks handle easy inputs, promising fast and low-energy inference, while larger networks operate on complex inputs to obtain the best performance. In an RL environment with sparse and delayed rewards, there is an abundance of `easy' states which can be exploited to obtain inference speedup with low compute effort.

For eg., consider a task where an RL agent is supposed to learn vision-based navigation in an environment full of obstacles. For an input state where the agent is surrounded by multiple obstacles, the image frame will contain more complex features. The predicted action will be of critical importance to avoid an impending collision (`hard' state). In contrast, for input states with obstacles far away from the agent, comparatively simpler features need to be identified to predict the optimal action (`easy' state). 

To that effect, we propose a reconfigurable architecture with preemptive-exits for efficient deep-reinforcement learning (RAPID-RL). We present our approach using Deep Q-learning (DQN), a model-free RL algorithm, but it can be generalized to any other RL algorithm employing DNNs. Our work aims to lessen the inference energy and latency of deep RL agents by dynamically adjusting the compute effort based on the difficulty of input samples. We achieve this by appending the main neural network with additional branches to allow a fraction of the input samples to be inferred preemptively. Fig.~\ref{fig:arch} illustrates the RAPID-RL topology for a DNN comprising of three convolutional and two fully-connected layers with two side-branches. We exploit the fact that a DNN learns a hierarchy of features that transition from being generic to specific \cite{yosinski2015understanding} as we go deeper. Thus, a substantial fraction of the input states can be inferred correctly by only using the generic features without activating the latter layers. Such an approach was first proposed in \cite{panda2016conditional} to design a Conditional Deep Learning Network (CDLN) by adding early-exit branches to a DNN for image classification.

An active RL problem introduces additional challenges over standard DNN classification. Firstly, RL involves a dynamic replay memory comprised of experience collected during training, compared to a static dataset in standard DNN classification. Secondly, instead of having target action labels for every input state, an RL environment provides feedback in the form of sparse and delayed rewards. This demands a rework of methodology for deciding preemptive-exit at a branch. RAPID-RL extends the conditional-exit method to an active RL domain while addressing the above challenges. \\
In summary, we make the following contributions:
\begin{enumerate}
\item \textbf{Reconfigurable architecture with preemptive-exits:} We propose RAPID-RL, a framework that augments a standard DQN with multiple exit branches to allow a majority of input states to be inferred preemptively. This results in faster inference and reduction in the number of compute operations (OPS)\footnote{\label{x1} OPS is defined as the total number of multiply and accumulate operations during a forward pass in a neural network}.

\item \textbf{Sequential Q-learning:} We propose a sequential method for incrementally training the intermediate branches while ensuring optimal performance at each branch. This decouples the performance inter-dependency between branches and allows to construct a joint-DQN catering to a permissible power budget.

\item \textbf{Confidence score training:} We propose a strategy for learning the branch confidence scores in the obtained joint-DQN. This is used to decide preemptive exit at a branch during inference.
\end{enumerate}
The experimental results involve evaluation on four different Atari 2600 games (Pong, Space-Invaders, PacMan, and Breakout) and a drone navigation task in a simulated environment (PEDRA)~\cite{anwar2020tl}. We provide comparisons with a baseline-DQN (without preemptive-exit branches) and observe that our method provides a significant reduction in OPS while maintaining application performance.

\section{BACKGROUND \&  RELATED WORK} \label{sec:background}
\subsection{Deep Q-learning}\label{subsec:dqn}
Q-Learning~\cite{watkins1992q} is a traditional RL algorithm for finding the optimal policy ($\pi$) to earn maximal rewards in a given environment. In Deep Q-learning, a neural network is used as a function approximator to represent the agent policy ($\pi$). At every time instant $t$, the agent-environment interaction can be represented as a Markov Decision Process (MDP) with the tuple $<s, a, r, s'>$, where $s$  and $s'$ represent the current and next state of the environment, $a$ is the action taken by the agent, and $r$ is the obtained reward. 

All future rewards are discounted by a factor of $\gamma \in (0, 1)$ per time-step. The discounted future sum of rewards at time $t$ can be given by: $R_t = \sum\limits_{k=0}^{\infty}\gamma^k r_{t+k+1}$.
The optimal Q-function $Q^*(s, a)$ is defined as the maximal expected return ($R_t$) starting from state $s$, taking action $a$ and then following the policy $\pi$ given by:  $Q^*(s,a) = \max_{\pi} \mathbb{E}_\pi[R_t | s_t=s, a_t=a]$.
The goal of deep Q-learning is to approximate the optimal Q-function by iteratively applying the Bellman Equation \cite{bellman1954}. If the optimal Q function $Q^*(s', a')$ for state $s'$ in the next time-step is known for all possible actions $a'$, then:
\begin{equation}
    Q^*(s,a) = \mathbb{E}_\pi[r + \gamma \max_{a'} Q^*(s', a')]
\label{eq1}
\end{equation}

A deep Q-network is trained with a loss function as:
\begin{equation}
L = (r+\gamma \max_{a'} Q_{tgt}(s', a') - Q(s, a))^2
\label{eq2}
\end{equation}
where, $Q(s, a)$ is obtained from the online network being trained and $Q_{tgt}(s', a')$ from a target network, maintained as a copy of the online network updated periodically. 

\begin{figure*}[t]
\centering
\includegraphics[width=\linewidth, trim={0.0in, 0.15in, 0.2in, 0.0in}, clip]{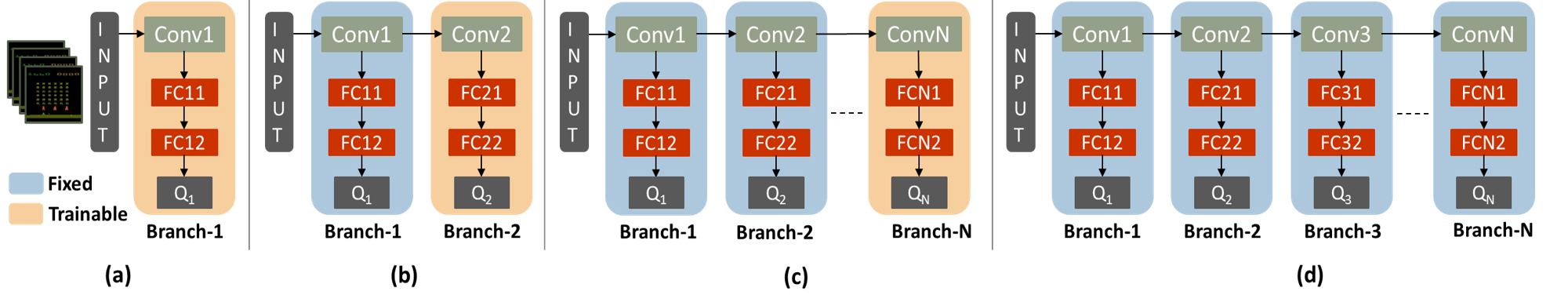}
\caption{\small{L2R sequential Q-learning in RAPID-RL. (a) Start with training the first branch (b) Subsequently add a new branch, fix the parameters on the already trained branch(es), and train the new branch. (c) Continue adding and training new branches until all branches are trained. (d) Finally, fix the parameters of all branches and construct the joint-DQN by selecting branches based on the target task.}}
\label{fig:training}
\vspace{-4mm}
\end{figure*}


DQN and its extensions have proven to be effective in gaming and control tasks and are the current go-to algorithms for RL problems. We employ the state-of-the-art DQN implementation named Rainbow DQN~\cite{hessel2017rainbow} in this work, which combines Double DQN~\cite{van2016deep}, Dueling networks~\cite{wang2015dueling}, Prioritized Replay~\cite{schaul2016prioritized}, Multi-step learning~\cite{suttonbarto2018}, Distributional RL~\cite{bellemare2017distributional} and Noisy nets~\cite{fortunato2018noisy} techniques.

\subsection{Related Work} \label{subsec:relatedwork}
Over the past few years, several optimization approaches, such as compression or pruning techniques \cite{han2015deep, han2015learning}, and binarized or low-precision network implementations \cite{courbariaux2016binarized, rastegari2016xnor} have been proposed to reduce the complexity of DNNs. An emerging data-driven category of optimization techniques \cite{panda2016conditional, panda2017energy, teerapittayanon2016branchynet, teerapittayanon2017distributed} performs early termination during inference to dynamically adjust the compute effort based on input by adding additional output branches. While the notion of RAPID-RL is derived from the latter approaches, it can be combined with compression or binarization techniques to improve the efficiency even further. 

It is noteworthy to mention that all prior works target static supervised image classification tasks with fixed output labels. Extending this approach to an RL setting requires several modifications without affecting the training convergence. RAPID-RL proposes a generic methodology for sequential training and inference in a multi-exit deep RL setting. The sequential training method allows for independent training of newly added branches. It also allows selecting a custom architecture during inference to keep the computational costs within the power budget of the target device.

\section{RAPID-RL FRAMEWORK} \label{sec:framework}
RAPID-RL architecture is an extension of a conventional DQN, with exit branches at intermediate layers. It utilizes Rainbow DQN~\cite{hessel2017rainbow}, an RL algorithm that integrates several advanced DQN methods for improved data utilization and algorithmic stability. The performance of RAPID-RL is compared to a baseline-DQN with no branches and trained using the same Rainbow DQN~\cite{hessel2017rainbow} algorithm.

The architectural design of RAPID-RL imposes several considerations, which can be treated as design hyper-parameters. These are (1) location of preemptive-exit points (side branches), (2) structure of the side branch (number of convolutional and fully-connected layers), (3) training routine for branches, and (4) preemptive-exit criteria for a branch. It is worth mentioning that each branch can have its own recursive branch structure (i.e. branch within a branch), resulting in a tree-like framework. For simplicity, we focus on an architecture with single-level branches added to the baseline-DQN without any nested and recursive pathways.

\subsection{RAPID-RL Training: L2R Sequential Q-learning} \label{subsec:training}
The baseline-DQN is trained by incorporating techniques from the Rainbow algorithm as discussed in Section-\ref{sec:background}A. DQN training involves exploration of the environment by using the action predictions from the online Q-network and storing the transitions in the experience replay. At regular training intervals, a batch of transitions is sampled from the current experience replay and evaluated using the online Q-network. This is followed by parameter updates based on the computed loss.

In RAPID-RL, a similar training procedure is followed, but in a sequential manner. Training starts from a shallow Q-network and progresses by incrementally adding and training new branches once all previous branches are trained. Whenever a new branch is added to the network, parameters of all previous branches are fixed and are not updated during training of the new branch. We term this training method as left-to-right (L2R) sequential training, since the network grows one branch at a time from left-to-right. Fig.~\ref{fig:training}(a,b,c) show the L2R sequential training method for a generic RAPID-RL framework with $N$ branches. The common layers are already trained to generate optimal feature maps. Thus, the performance of past branches remains unaffected when a new branch is trained. This leads to architecture with each branch having optimal performance.


Once all ($N$) branches are trained, a joint-DQN is constructed as shown in Fig.~\ref{fig:training}(d). The decoupling of training routines for each branch allows for a curated selection of branches to suit the needs of target edge applications. For eg., an L2R trained network consisting of four convolutional layers with branches at each convolutional layer can be re-built for an energy-constrained application with only the initial three convolutional layers and a branch at only the first convolutional layer. This does not require any further training of the selected branches.

\subsection{Branch Confidence training}
The obtained joint-DQN consists of optimized branches capable of delivering optimal performance individually. The latter branches perform better than the earlier branches as they operate on more specific feature maps. In order to obtain best tradeoff between energy and latency improvements while maintaining performance, an intelligent branch-exit selection strategy need to be devised. The strategy needs to keep into account the dynamic nature of the dataset, feedback only in the form of delayed rewards and the varying difficulty levels of the input states.

\begin{figure*}[t]
\centering
\includegraphics[width=\linewidth]{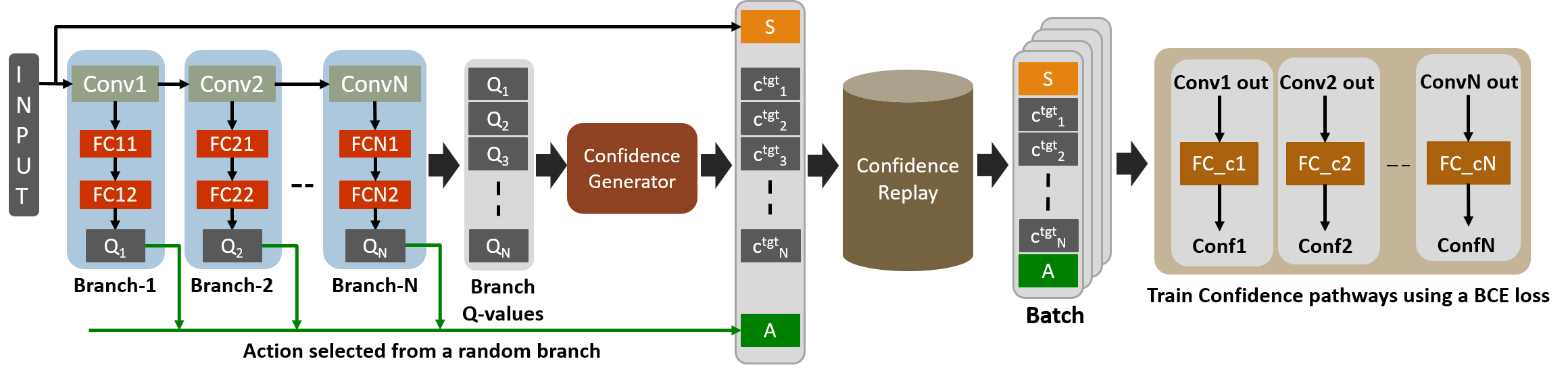}
\caption{\small{Confidence replay generation using Q-values obtained from all branches of a trained joint-DQN. The generated confidence replay is then used for training the confidence pathways at all the branches.}}
\label{fig:conf_training}
\vspace{-3mm}
\end{figure*}

To address these, we propose to augment each branch in the joint-DQN with a confidence pathway that has the same architecture as the branch classifier (fully-connected layers) but generates a single sigmoid output denoting the confidence score for that branch. Training the confidence pathways in an RL setting involves two parallel processes as below:

\subsubsection{\textbf{Confidence replay generation}} A confidence replay is generated to store the target confidence labels for training the confidence pathways. Environment exploration is carried out, as usual, however, instead of state-action-reward pairs, the experience now contains the state, action, and the target confidence labels corresponding to each branch $<s, a, c^{tgt}_1, c^{tgt}_2, c^{tgt}_3, ... c^{tgt}_N>$. The action ($a$) to be taken for a given input state during exploration is randomly chosen from the actions predicted by the branches ($a_i$) to prohibit the experience replay from becoming biased towards a particular branch. Let $Q_i(s)$ represent the Q-values of branch $i$ for a given state $s$ and for all possible actions. Then, the confidence generator works as follows:
\begin{equation}
    a_i = \argmax_a Q_i(s,a)  \quad \forall i\in\{1, 2, \ldots N\}
\end{equation}
\begin{equation}
    a = random(a_1, a_2, a_3, ... a_N)
\end{equation}

The target confidence labels ($c^{tgt}_i(s)$) for each branch are generated by comparing the Q-values of the predicted actions (maximum Q-value across all actions) for each branch ($Q^{max}_i(s)$) with the maximum Q-value ($Q^{max}(s)$) across all $N$ branches as below: 
\begin{equation}
    Q^{max}_i(s) = \max_a Q_i(s,a) \quad \forall i\in\{1, 2, \ldots N\} 
\end{equation}
\begin{equation}
    Q^{max}(s) = \max_i(Q^{max}_i(s)) 
\end{equation}
\begin{equation}
    c^{tgt}_i(s) = \begin{cases}
           1, & \text{if $Q^{max}_i(s) > C*Q^{max}(s)$}, \\
           0, & \text{otherwise}
          \end{cases}
\end{equation}
where $C\in(0,1]$, is the confidence acceptance threshold. A high $C$ puts emphasis on high performance while a low $C$ leads to frequent preemptive-exits at the cost of performance.

\begin{figure}[b]
\centering
\includegraphics[width=\linewidth, trim={0.75in, 0.75in, 0.75in, 0.75in}, clip]{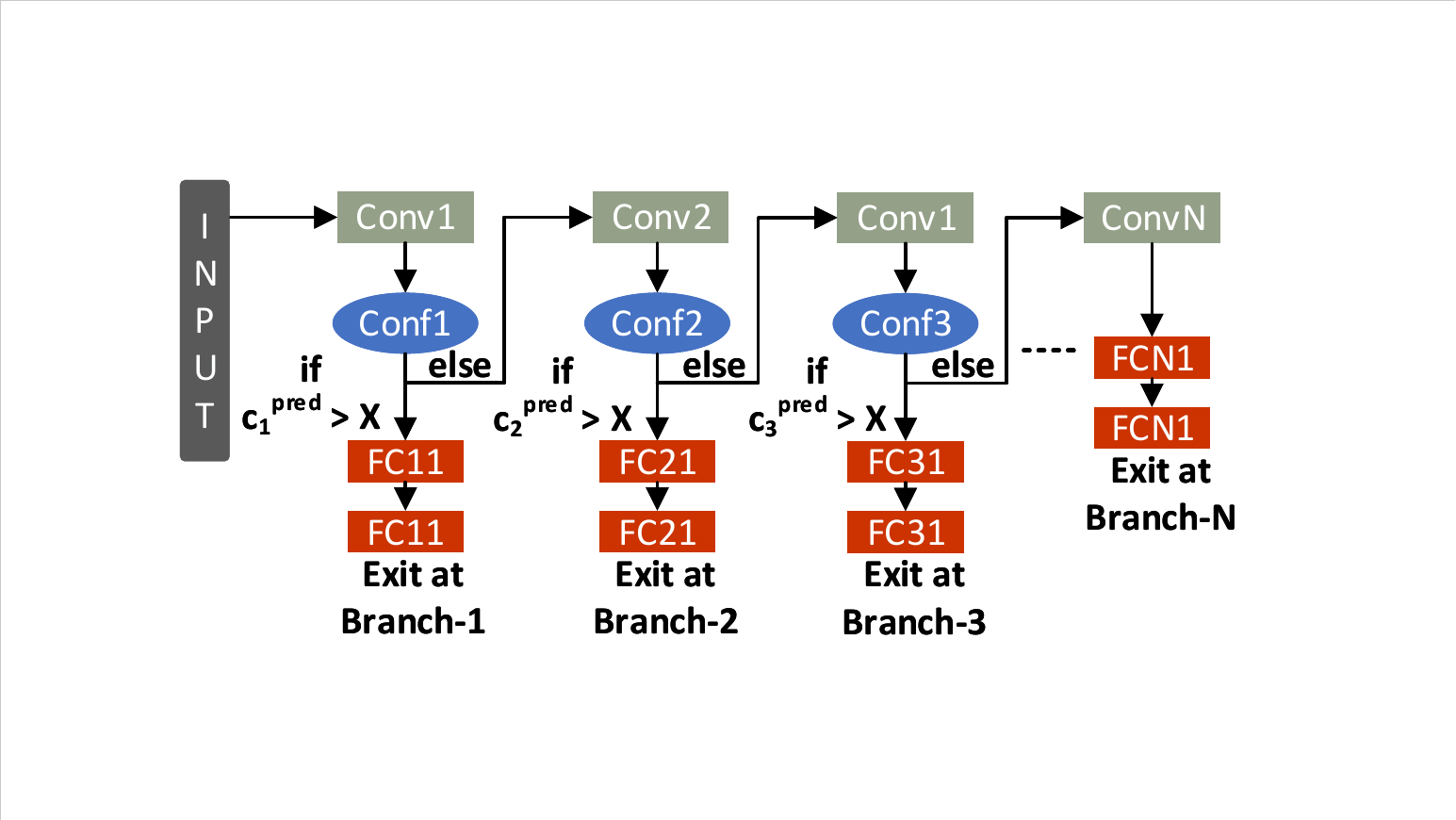}
\caption{\small{Preemptive-exit-based inference in the RAPID-RL.}} 
\label{fig:infer}
\end{figure}

\subsubsection{\textbf{Confidence score training}}
At regular training intervals, a batch of transitions is sampled from the current confidence replay and used to train all the confidence pathways together. A binary cross-entropy (BCE) loss is computed between the predicted confidence output ($c^{pred}_i(s)$) and target confidence labels ($c^{tgt}_i(s)$) from the confidence replay, $L_{c_i} = BCE(c^{pred}_i, c^{tgt}_i)$. The total mean loss is computed as:
\begin{equation}
L_{c} = \frac{1}{N}\sum_{i=1}^{N} L_{c_i}
\end{equation}

\subsection{RAPID-RL Inference - Preemptive-exit}
After training all the branches and the confidence pathways, the intermediate branches can be used as exit points to infer a decision preemptively without activating the full network hierarchy.
Fig.~\ref{fig:infer} shows the preemptive-exit-based inference procedure in the RAPID-RL framework. If the confidence score ($c^{pred}_i$) for a given state at an intermediate branch ($i$) is greater than a pre-specified preemptive-exit threshold ($X \in (0, 1)$), then the Q-value for that branch is considered `good' enough to decide the action to be performed by the agent(`easy' state). The inference is thus terminated at that branch. However, if the confidence score ($c^{pred}_i$) is lower than $X$, the side-branch is deemed not confident, and the given state is assessed to be a `hard' state. The process continues to the next branch exit until a confidence score greater than $X$ is obtained or the last branch is reached . In the extreme case when $X=0$, all states are preemptively handled at the first branch, leading to the highest benefits in energy and latency. This however would lead to poor rewards. On the other hand, for $X=1$, majority of the states are handled at the last ($N^{th}$) branch leading to high rewards, but low benefits in energy and latency. Note, that preemptive-exit at the $k^{th}$ branch would involve evaluating the confidence pathways of all past $k$-$1$ branches. This leads to a minor overhead in terms of computational cost when compared to evaluating the $k^{th}$ branch directly. Our results show that the benefits of being able to intelligently choose a preemptive-exit while maintaining a high long term reward largely overshadows this minor computational cost.
\begin{figure*}[t]
\centering
\includegraphics[width=\linewidth, trim={2.0in, 1.2in, 2.0in, 0.75in}, clip]{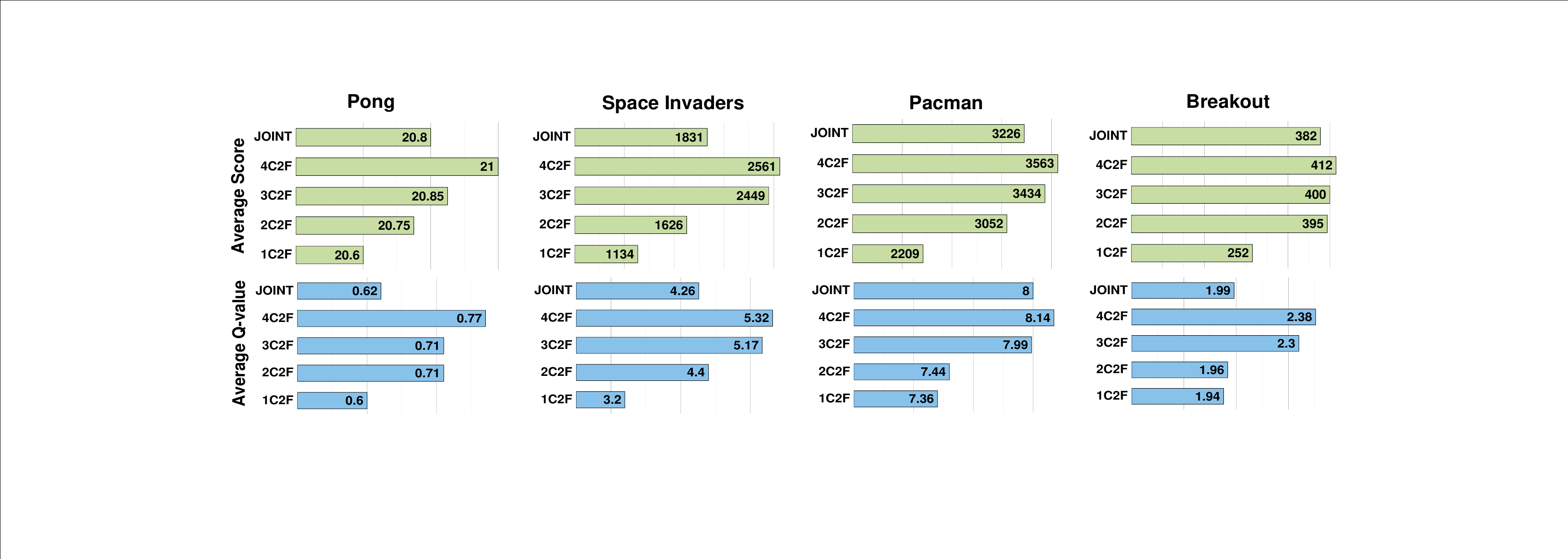}
\caption{\small{Average score and Q-value for different branches as well as the joint-DQN for Atari 2600 games.}}
\label{fig:score_atari}
\vspace{-3mm}
\end{figure*}
\section{EXPERIMENTAL RESULTS}
\subsection{Experimental Setup}
Table-\ref{tab:common} lists the common hyperparameters for both tasks corresponding to the Rainbow DQN algorithm.

\begin{table}[ht]
\vspace{-2mm}
\centering
\caption{Common hyperparameters.}
\begin{tabular}{|c|c|p{\textwidth}} 
\hline \vspace{-2.8mm}\\
\textbf{Hyperparameter} & \textbf{Value}  \\
\hline
Training Steps (T) & $20 \times 10^6$\\
\hline
Batch Size (B) & 32\\
\hline
Replay Memory size & $10^5$\\
\hline
Gamma ($\gamma$) & 0.99\\
\hline
Learning start & $80 \times 10^3$\\
\hline
Target Update frequency & $8 \times 10^3$\\
\hline
Learning Rate & $6.25 \times 10^5$\\
\hline
Adam Epsilon ($\epsilon$) & $1.5 \times 10^-4$\\
\hline
Train interval & $4$\\
\hline
Validation Episodes & $20$\\
\hline
Validation interval & $1 \times 10^6$\\
\hline
Priority weight & $0.4$\\
\hline
Priority exponent & $0.5$\\
\hline
Multi-step learning parameter & $3$\\
\hline
Number of atoms & $51$\\
\hline
Noisy Net \{$V_{min}$, $V_{max}$, $\sigma_0$\}  & ${-10, 10, 0.5}$\\
\hline
Confidence threshold (C) & $0.8$\\
\hline
Preemptive threshold (P) & $0.7$\\
\hline
\end{tabular}
\vspace{-2mm}
\label{tab:common}
\end{table}

\begin{table}[ht]
\caption{Architecture for the joint-DQN}
\vspace{-2mm}
\begin{center}
\resizebox{0.49\textwidth}{!}{
\begin{tabular}{ccccc} 
 
    \multicolumn{5}{c}{\textbf{Convolutional Layers}} \\ \cline{1-5}
        Layer &  Kernel Size & Stride & In Channels & Out Channels \\ \cline{1-5} \vspace{-2.5mm}\\
        Conv1 & 8$\times$8 & 4 & 4 & 16 \\
        Conv2 & 4$\times$4 & 2 & 16 & 32 \\
        Conv3 & 3$\times$3 & 1 & 32 & 64 \\
        Conv4 & 3$\times$3 & 1 & 64 & 128 \\ \cline{1-5} \vspace{-1.5mm}
    \end{tabular}
}
\resizebox{0.49\textwidth}{!}{
\begin{tabular}{cccccccc}
  \multicolumn{3}{c}{\textbf{Branch-1 at Conv1}} &  & \multicolumn{3}{c}{\textbf{Branch-2 at Conv2}} \\ \cline{1-3} \cline{5-7} \vspace{-2.5mm}\\ 
Layer & Input & Output &  & Layer & Input & Output \\ \cline{1-3}  \cline{5-7} \vspace{-2.5mm} \\
AvgPool1 & 16 & 16 &  & AvgPool2 & 32 & 32 \\
FC11 & 400 & 256 &  & FC21 & 800 & 256 \\
FC12 & 256 & $n_{actions}$ &  & FC22 & 256 & $n_{actions}$ \\ \cline{1-3}  \cline{5-7} \vspace{-1.5mm}
\end{tabular}}

\resizebox{0.49\textwidth}{!}{
\begin{tabular}{cccccccc}
  \multicolumn{3}{c}{\textbf{Branch-3 at Conv3}} &  & \multicolumn{3}{c}{\textbf{ Branch-4 (Main) at Conv4}} \\ \cline{1-3} \cline{5-7} \vspace{-2.5mm}\\ 
Layer & Input & Output &  & Layer & Input & Output \\ \cline{1-3}  \cline{5-7} \vspace{-2.5mm} \\
AvgPool3 & 64 & 64 &  & AvgPool4 & 128 & 128 \\
FC31 & 1600 & 256 &  & FC41 & 3200 & 256 \\
FC32 & 256 & $n_{actions}$ &  & FC42 & 256 & $n_{actions}$ \\ \cline{1-3}  \cline{5-7} \vspace{-3.8mm}
\end{tabular}}

\end{center}
\begin{flushright}\small{\;* $n_{actions}$ refers to task specific total number of possible actions}\end{flushright}
\vspace{-8mm}
\label{tab:net}
\end{table}



\subsubsection{Network Architecture}
For all the experiments, we employ a DNN architecture with four convolutional and two fully-connected layers (4C2F) for the baseline-DQN. This architecture is a modification over the standard architecture used in~\cite{hessel2017rainbow}. The joint-DQN contains the 4C2F path as its main branch with intermediate branch exits at all previous convolutional layers. Each branch exit consists of a classifier with two fully-connected layers leading to 1C2F, 2C2F, and 3C2F branches. Table-\ref{tab:net} depicts the joint-DQN architecture. The feature maps acting as inputs to the branch classifiers are kept identical in size by using an adaptive average pooling layer with an output size of $5\times5$. This is done to maintain similar classification ability and prevent the explosion of parameters in the fully-connected branch classifiers. 


\subsubsection{Evaluation Metrics}
The efficacy of the joint-DQN is analyzed in terms of its performance and efficiency. The performance is represented by the average score (reward) for inference over several episodes. The energy and latency benefits are realized due to the reduction in the number of operations (OPS) performed by the joint-DQN on average considering preemptive-exits. The performance of the baseline-DQN is generally better than the joint-DQN as it employs the deepest network for every input state. The performance ratio ($P$) is obtained as the ratio between the average score obtained by the joint-DQN and the baseline-DQN. The ratio between the OPS incurred ($E$) during a forward pass by the joint-DQN and the baseline-DQN represents the energy and latency improvements. 
Note, that the OPS ratio ($E$) corresponding to the $k^{\text{th}}$ branch is computed taking into account the OPS incurred during a forward pass through the confidence pathways at all previous $(k-1)^{\text{th}}$ branches.

\subsection{Atari 2600}
\begin{figure*}[t]
\centering
\includegraphics[width=\linewidth, trim={2in, 0.75in, 1.7in, 1in}, clip]{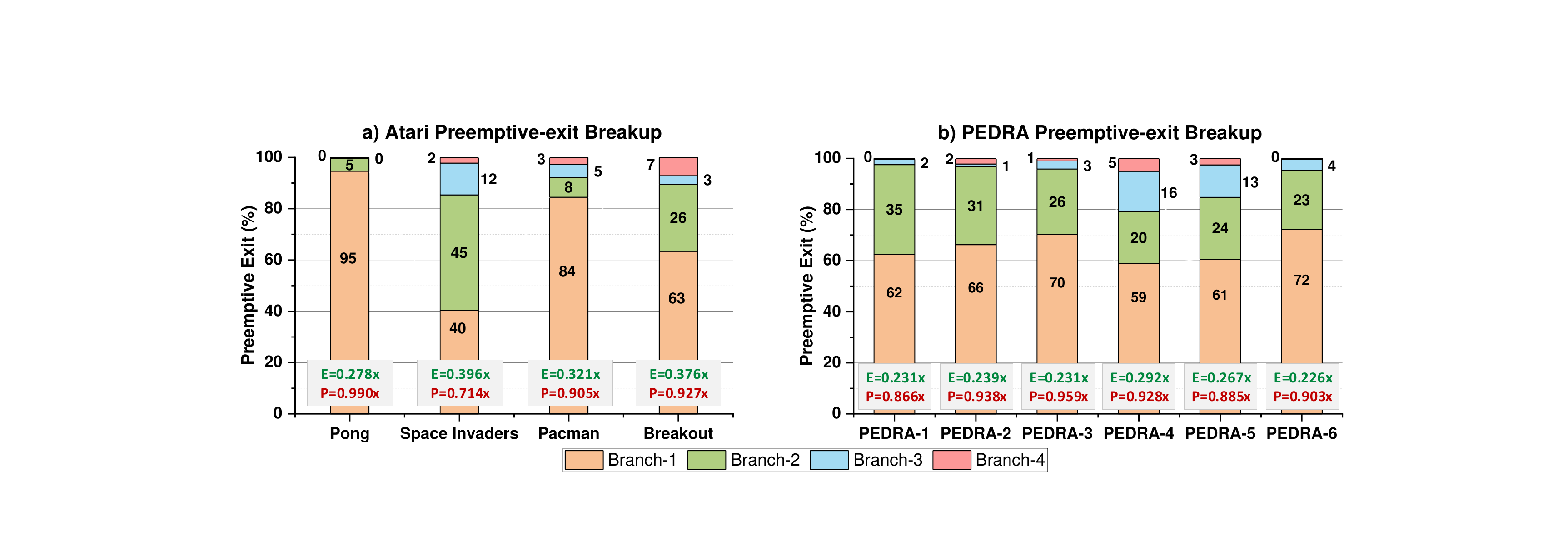}
\caption{\small{Percentage breakup for preemptive-exit along with OPS (E) and Performance (P) ratios compared to baseline-DQN. (a) For Atari games (Pong, Space-Invaders, PacMan and Breakout, (b) For $6$ different starting positions in the PEDRA Indoor-long environment.}}
\label{fig:results}
\vspace{-3mm}
\end{figure*}

We evaluate RAPID-RL on Atari 2600 games - Pong, Space-Invaders, PacMan, and Breakout in the Arcade Learning Environment (ALE)~\cite{bellemare2015ALE}. The input frames are pre-processed to $84\times84$ size and $4$ such consecutive frames are passed as input to the network. The average score and average Q-values obtained by the different branch networks independently, as well as by the joint-DQN are shown in Fig.~\ref{fig:score_atari}. Fig.~\ref{fig:results}(a) shows the benefits of the joint-DQN based on the percentage of states preemptively exiting at each branch over several inference episodes.

It was observed that for a simple game such as Pong, even the 1C2F branch was able to obtain near identical average score as the main branch (4C2F) leading to a ($\sim95\%$) preemptive-exit. On the other hand, for Space Invaders, which is a more difficult task, the 1C2F branch independently performs poorly producing less than half the score ($1134$) compared to the 4C2F branch ($2561$). This highlights the presence of `hard' states and leads to only $40\%$ preemptive-exit at the 1C2F branch and a corresponding OPS ratio of $0.39\times$. In such a scenario, the joint-DQN can be reconfigured to remove the 1C2F branch and have intermediate branches at only $Conv2$ and $Conv3$. This new configuration leads to a performance ratio of $0.86\times$ with a $7\%$ increase in OPS and $\sim82\%$ exit at the 2C2F branch. PacMan and Breakout show much better results with $\sim85\%$ and $\sim63\%$ preemptive-exit at the 1C2F branch respectively while maintaining a performance ratio greater than $0.9\times$. The OPS ratio on average over all four Atari games is $0.34\times$.

\subsection{Drone Autonomous Navigation - PEDRA}
\begin{figure}[h]
\vspace{-2mm}
\centering
\includegraphics[width=\linewidth]{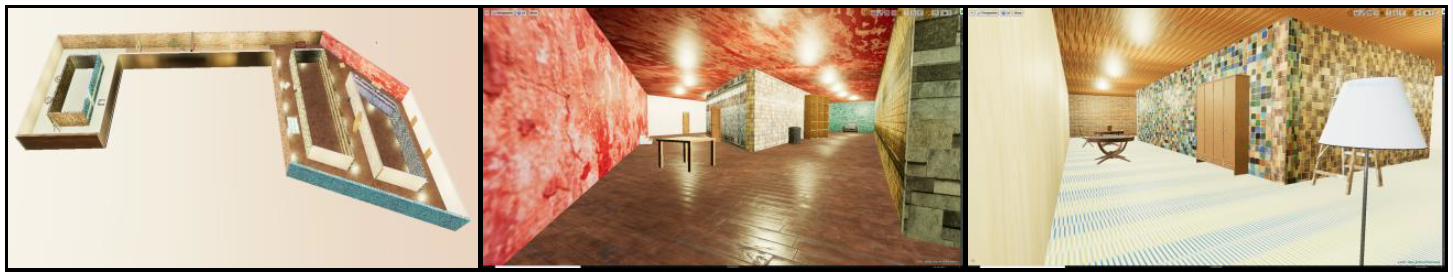}
\caption{\small{Floor-plan of the Indoor-long environment.}}
\label{fig:floorplan}
\vspace{-4mm}
\end{figure}

We also perform experiments on a more real-world-like task of drone autonomous navigation on PEDRA~\cite{yoon2019hierarchical, anwar2020tl}. PEDRA is an open-source drone autonomous simulator based on the Unreal gaming engine \cite{unreal} and  AirSim~\cite{airsim2017fsr}. The objective of this task is for the drone to fly as long as possible without colliding with obstacles in the environment. The Indoor-long environment (shown in Fig.~\ref{fig:floorplan} from PEDRA is used for our experiments. The input to the DQN is an RGB frame of size $227\times227$ captured from the front-facing camera of the drone and the output is a $5\times5$ array of Q-values, each corresponding to one action in a grid action space. An action in the $5\times5$ grid corresponds to the direction (pitch and yaw) in which the drone will move by a fixed $0.5m$. The performance metric used is the Mean Safe Flight distance (MSF) defined as the distance traveled by the drone (in $m$) starting from a given initial position before colliding with an obstacle, averaged over several episodes.

\begin{table}[h]
    \centering
    \caption{Mean Safe Flight (MSF) distance for different branches averaged over all initial positions.}
    \begin{tabular}{cccccc} 
    \hline \vspace{-2.5mm}\\ \vspace{1mm}
        Metric &  1C2F & 2C2F & 3C2F & 4C2F & Joint \\ \hline \vspace{-2.8mm} \\
        Mean Safe Flight (m) & 504.2 & 642.8 & 786.2 & 894.5 & 811.6  \\ \vspace{-2.8mm} \\ \hline \\
    \end{tabular}
    \label{tab:pedra}
    \vspace{-4mm}
\end{table}

The training was carried out while cycling between the $6$ different initial positions, switching after every $50K$ steps. Inference involved evaluating the MSF over each of the $6$ initial positions, with $20$ episodes for each position. Table-\ref{tab:pedra} provides the MSF obtained by the different branches as well as the joint-DQN averaged over all $6$ initial positions. Fig.~\ref{fig:results}(b) presents the preemptive-exit breakup for inference starting at the $6$ initial positions, highlighting the corresponding OPS ($E$) and performance ($P$) ratios. We observe that the network preemptively exits at the 1C2F branch for $\sim65\%$ of the states while having a performance ratio of 0.91$\times$ and an OPS ratio of $0.25\times$ averaged over all initial positions. Compared to Atari games, the later convolutional layers in this task incur significantly more operations due to larger feature maps. This leads to a lower OPS ratio for identical percentage preemptive-exit at an earlier layer in this task to have a lower OPS ratio compared to the Atari games.

\section{ABLATION STUDY}
We perform an ablation study to analyze the effect of variation in training methodology on the overall performance of the constructed joint-DQN. In addition to our proposed methodology of L2R (left-to-right) sequential training, we investigate two other training methodologies as follows:

\textbf{R2L training}: In the R2L (right-to-left) method, sequential training is performed starting with the main branch and subsequently adding and training additional branches at previous convolutional layers. This method although could achieve a similar score as the L2R network when evaluating through its main branch, it suffered heavily in terms of individual branch performances. It also sacrificed configurability, as adding a new branch at the end of the network required retraining all the previous branches again.

\textbf{Coupled training}: This method fixates the branch structure of the joint-DQN. All branches are trained in a coupled fashion using a single combined loss. This method led to faster training, but offered no configurability and led to poor performance at the intermediate branches. The training was performed for 40M frames due to more parameters. 

Table-\ref{tab:train_ablation} provides training ablations for Pong and Space Invaders and highlights the benefits of the L2R approach.

\begin{table}[t]
\caption{Training ablation on Pong and Space Invaders}
\vspace{-2mm}
\begin{center}
\resizebox{0.49\textwidth}{!}{
\begin{tabular}{llcccccccccc}
\hline
  & & \multicolumn{4}{c}{Pong} &  & \multicolumn{4}{c}{Space Invaders} \\ \cline{3-7} \cline{8-11} \vspace{-2.5mm}\\
 Method & & 1C2F & 2C2F & 3C2F & 4C2F &  & 1C2F & 2C2F & 3C2F & 4C2F \\ \cline{1-1}  \cline{3-6} \cline{8-11} \vspace{-2.5mm} \\
L2R            &  &  \textbf{20.6}    & \textbf{20.75}    & \textbf{20.85}    & \textbf{21.0}    &  & \textbf{1134}    & \textbf{1626}    & \textbf{2449}    & \textbf{2561}     \\ \vspace{-3.5mm} \\
R2L              & &  1.03     & 8.65    & 19.3    & 21.0     &  & 628     & 1293    & 2242    & 2617    \\ \vspace{-3.5mm} \\
Coupled              & &  -0.4     & 1.8    & 10.6    & 20.3     &  & 387     & 948    & 864    & 915   \\ \hline \vspace{-3.5mm} 
\end{tabular}}
\end{center}
\vspace{-7.5mm}
\label{tab:train_ablation}
\end{table}

\section{CONCLUSION}
We proposed RAPID-RL, an RL framework to perform training and inference with networks containing preemptive exits. RAPID-RL identifies and addresses the challenges posed by active RL environments compared to standard DNN classification. We presented sequential Q-learning to construct a joint-DQN as well as a method to train the confidence pathways in RAPID-RL. We obtain average energy ratios of $0.34\times$ ($0.25\times$) and corresponding performance ratios of $0.88\times$ ($0.91\times$) on Atari (Drone navigation) tasks respectively, compared to corresponding baseline-DQN architectures. The preemptive-exit capability and reconfigurability offered by RAPID-RL allows for fast and efficient implementations on resource-constrained edge devices while maintaining overall performance of the RL agent.

\bibliography{main.bbl} 
\bibliographystyle{ieeetr}

\end{document}